\documentclass[runningheads]{llncs}
\usepackage[T1]{fontenc}

\usepackage{amsmath}
\usepackage{amsfonts}
\usepackage{dsfont}
\usepackage{graphicx}
\usepackage{tikz}
\usepackage{multirow}
\usepackage{array}
\usepackage{tabularx,booktabs}
\usepackage{subcaption}
\usepackage{dirtytalk}
\usepackage{bm}
\usepackage{dsfont}
\usepackage{float}
\usetikzlibrary{arrows.meta}
\usepackage{color}
\usepackage{hyperref}
\usepackage{comment}
\usepackage[misc]{ifsym}

\newcommand{\repeatthanks}{\textsuperscript{\thefootnote}}

\begin{document}

\title{Evaluating Visual Explanations of Attention Maps for Transformer-based Medical Imaging}

\titlerunning{Evaluating Visual Explanations of Attention Maps}

\author{Minjae Chung\inst{1}\thanks{Equal contribution.}  \and
Jong Bum Won\inst{1}\repeatthanks\and
Ganghyun Kim\inst{1}\repeatthanks\and\\
Yujin Kim\inst{1}\repeatthanks\and 
Utku Ozbulak\inst{1,2}(\Letter)}

\authorrunning{Chung et al.}

\institute{Center for Biosystems and Biotech Data Science,\\ Ghent University Global Campus, Republic of Korea \and
Department of Electronics and Information Systems, Ghent University, Belgium\\
(\Letter) \email{utku.ozbulak@ghent.ac.kr}
}
%
%
\maketitle              
\begin{abstract}
Although Vision Transformers (ViTs) have recently demonstrated superior performance in medical imaging problems, they face explainability issues similar to previous architectures such as convolutional neural networks. Recent research efforts suggest that attention maps, which are part of decision-making process of ViTs can potentially address the explainability issue by identifying regions influencing predictions, especially in models pretrained with self-supervised learning. In this work, we compare the visual explanations of attention maps to other commonly used methods for medical imaging problems. To do so, we employ four distinct medical imaging datasets that involve the identification of (1) colonic polyps, (2) breast tumors, (3) esophageal inflammation, and (4) bone fractures and hardware implants. Through large-scale experiments on the aforementioned datasets using various supervised and self-supervised pretrained ViTs, we find that although attention maps show promise under certain conditions and generally surpass GradCAM in explainability, they are outperformed by transformer-specific interpretability methods. Our findings indicate that the efficacy of attention maps as a method of interpretability is context-dependent and may be limited as they do not consistently provide the comprehensive insights required for robust medical decision-making.

\keywords{Interpretability \and Vision Transformers \and Attention Maps}
\end{abstract}

\section{Introduction}
The global population is experiencing a significant shift as advancements in modern medicine have led to increased life expectancy and a rising median age which necessitates continuous medical attention, particularly for the elderly~\cite{US_med}. Recent shortage of medical workforce in many countries exacerbated these challenges, making it difficult for healthcare systems to meet the growing demand~\cite{korea_med,US_med}. 

An innovative solution to address the issues faced by global healthcare is the use of artificial intelligence (AI) for accurate and fast diagnosis, especially for medical imaging problems. Nowadays, such solutions often utilize the recently discovered Vision Transformer (ViT) architecture, which originates from transformers which make use of attention mechanism. As it stands, ViTs achieve state-of-the-art results in several medical imaging tasks and have the potential to be become the dominant architecture in the field, possibly replacing convolutional neural networks (CNNs) in near future~\cite{matsoukas2021time,shamshad2023transformers,singhal2023towards}.

Despite notable advancements in the accuracy of ViT-based medical imaging models, these models face challenges regarding explainability similar to CNNs. These explainability issues are one of the major obstacles preventing the large-scale adoption of AI-based diagnostic methods in medical institutes~\cite{hatherley2020limits,rajpurkar2022ai,kerasidou2022before}. To enhance the trustworthiness of predictions made by ViTs and to address the aforementioned interpretability challenges, numerous methods have been proposed in the literature~\cite{vis_survey1,vis_survey3}. Among these methods, attention maps stand out as an explainability technique, since they are part of the ViT architecture and directly influence the predictions made by the model. However, there is an ongoing debate~\cite{debate} on the usage of attention maps as explanation, with some research efforts advocating for~\cite{dino,darcet2023vision,att_cv_vs_nlp} and some against it~\cite{attention_not_interpretable,is_att_interpretable}. Complicating this issue further, recent research on self-supervised learning (SSL) for ViTs has shown that SSL pretraining improves the efficacy of attention maps for explainability~\cite{dino}. However, these results are often demonstrated in the context of natural images, and it remains unclear if these findings apply to medical imaging problems with specific modalities.

In this work, we evaluate the suitability of attention maps for interpreting transformer-based medical imaging models by comparing them with other commonly used methods for explainability. Our study complements that of~\cite{komorowski2023towards} by expanding the scope to include a wider range of datasets, including non-radiology medical images. Additionally, we also utilize state-of-the-art SSL pretrained ViTs to investigate the impact of self-supervised pretraining on explainability. 

\section{Experimental Setup}

\begin{figure}[t!]
\centering
\begin{subfigure}{0.244\textwidth}
\centering
\includegraphics[width=0.48\textwidth]{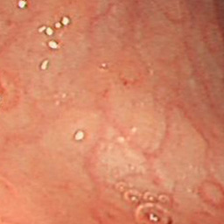}
\includegraphics[width=0.48\textwidth]{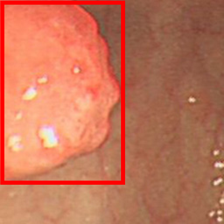}
\caption{CP-Child}
\label{fig:cpchild_ims}
\end{subfigure}
\begin{subfigure}{0.244\textwidth}
\centering
\includegraphics[width=0.48\textwidth]{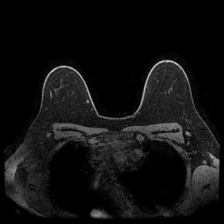}
\includegraphics[width=0.48\textwidth]{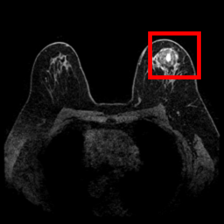}
\caption{DUKE}
\label{fig:duke_ims}
\end{subfigure}
\begin{subfigure}{0.244\textwidth}
\centering
\includegraphics[width=0.48\textwidth]{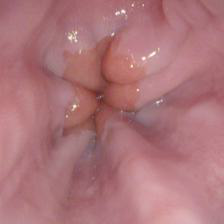}
\includegraphics[width=0.48\textwidth]{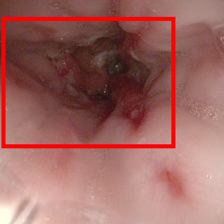}
\caption{Kvasir}
\label{fig:kvasir_ims}
\end{subfigure}
\begin{subfigure}{0.244\textwidth}
\centering
\includegraphics[width=0.48\textwidth]{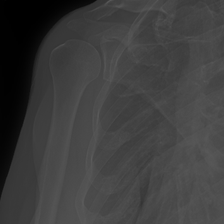}
\includegraphics[width=0.48\textwidth]{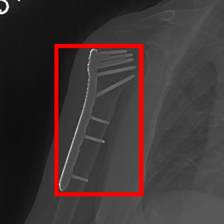}
\caption{MURA}
\label{fig:mura_ims}
\end{subfigure}
\caption{Examples of medical images in (a) CP-Child, (b) DUKE, (c) Kvasir, and (d) MURA datasets. The left images shows benign (disease-negative) images, while the right shows malignant (disease-positive) images for each dataset. Red annotation boxes highlight the regions with diseases.}

\label{fig:dataset_examples}
\end{figure}

\begin{table}[t!]
\centering
\scriptsize
\caption{Details for datasets used in this study, including the number of negative and positive samples, and the classification tasks, are provided.}
\label{tbl:dataset}
\begin{tabular}{cccccc}
\toprule
\multirow{3}{*}{Dataset} & \multicolumn{2}{c}{Training} & \multicolumn{2}{c}{Validation} & \multirow{3}{*}{Classification task} \\
\cmidrule[0.75pt]{2-5}
~ & \phantom{--}Negative\phantom{--} & \phantom{--}Positive\phantom{--} & \phantom{--}Negative\phantom{--} & \phantom{--}Positive & ~ \\
\midrule
CP-Child & 7,000 & 1,100 & 1,100 & 300 & Colonic polyp \\
DUKE & 17,642 & 17,546 & 4,350 & 4,446 & Breast tumor \\
Kvasir & 800 & 800 & 200 & 200 & Gastrointestinal disease \\
MURA & 21,935 & 14,873  & 1,667 & 1,530 & Bone abnormality \\
\bottomrule
\end{tabular}
\end{table}

\subsection{Datasets}
\label{sec:dataset}

To conduct a comprehensive investigation widely applicable to medical imaging, we selected four distinct datasets differing in image modality and body part. Table~\ref{tbl:dataset} contains the task and the number of images in the training and validation sets for each dataset. Below, we briefly describe each dataset.

\textbf{CP-Child Colonic Polyp Detection Dataset}\,\textendash\,Obtained from 1,600 patients, the task of this dataset is the detection of colonic polyps in the gastrointestinal track~\cite{wang2020improved}.

\textbf{DUKE Breast Cancer Dataset}\,\textendash\,This dataset, obtained from 922 patients with invasive breast cancers collected at Duke Hospital over 14 years, is one of the largest publicly available breast cancer 3D MRI image datasets, where the task is identifying the presence of breast tumors~\cite{duke_dataset}.

\textbf{Kvasir Gastrointestinal Disease Detection Dataset}\,\textendash\,Part of the Kvasir dataset~\cite{kvasir}, we utilize the Z-line subcategory, where the task is to distinguish between a healthy Z-line and esophagitis at the junction between the esophagus and stomach.

\textbf{MURA Musculoskeletal Radiograph Dataset}\,\textendash\,Obtained from 14,863 bone X-ray studies, Stanford's MURA dataset involves the identification of abnormalities such as bone fractures and hardware implants~\cite{mura}.

\subsection{Models}
\label{sec:models}

To evaluate interpretability methods, we employ widely used Vision Transformer-Base/16 (ViT-B/16)~\cite{vit}. This model processes images of size $224 \times 224$ with $16 \times 16$ image tokens, resulting in total $196$ tokens. We modify the final linear layer for our two-class classification problems and use four ViT-B/16 models with different weight initializations: random (i.e., from scratch), supervised pretrained, discriminatively pretrained using DINO~\cite{dino}, and generatively pretrained using MAE~\cite{mae}. Both supervised and self-supervised pretrained models are pretrained using the ImageNet dataset~\cite{ILSVRC15:rus}.

\begin{figure}[t!]
\centering
\centering
\includegraphics[width=1.0\textwidth]{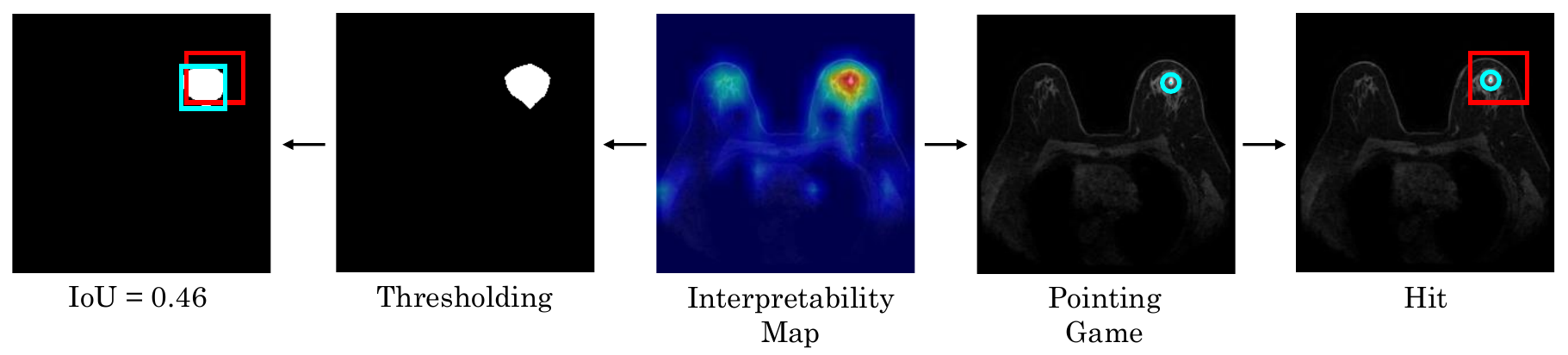}
\caption{Evaluation of interpretability maps is visualized. The right side illustrates the pointing game, identifying the most significant point and checking for a hit against the annotation box (red). The left part illustrates the IoU overlap, including the thresholding opeartion to create a binary mask and the calculation of IoU.}
\label{fig:vis_example}
\end{figure}

\subsection{Interpretability Methods}
\label{sec:int_methods}

To evaluate the efficacy of attention maps and to compare it with other methods, we use GradCAM~\cite{vis_grad_cam} and the transformer-specific interpretability method proposed in~\cite{chefer2021transformer} (henceforth referred to as the \say{Chefer method} in the rest of the paper).

\textbf{GradCAM}\,\textendash\,
GradCAM, an extension of Class Activation Mapping (CAM), ensures interpretability for CNNs by incorporating both forward activations as well as gradients originating from the selected class~\cite{vis_cam,vis_grad_cam}. Despite various extensions, GradCAM remains widely adopted for its robust results~\cite{vis_grad_cam_plusplus}. Following the past research efforts, we employ GradCAM on the final self-attention layer of ViT-B/16~\cite{chefer2021transformer}.

\textbf{Attention Maps}\,\textendash\,
In the ViT-B/16 setup, images are tokenized and prepended with an additional \texttt{[cls]} token, resulting in 197 tokens~\cite{vit}. The \texttt{[cls]} token, in conjunction with image patch tokens, are utilized to form attention maps for the self-attention mechanism. The \texttt{[cls]} token is presumed to encapsulate key information about the contributions of specific image patches to the model’s classification decisions. Following previous research, we investigate attention map of the \texttt{[cls]} token at the final layer~\cite{dino,vit,ozbulak2023know}.

\textbf{Chefer Method}\,\textendash\,
\cite{chefer2021transformer} argue that attention maps alone are insufficient for transformer interpretability, focusing narrowly on attention scores obtained from the inner products of queries and keys, and discarding other components. To address these limitations,~\cite{chefer2021transformer} propose a method that propagates gradient and relevance scores backward through all layers from the classification output.

\subsection{Interpretability evaluation}
\label{sec:evaluation}

In order to enable a fair comparison across explainability methods, for each dataset, 50 disease-positive images from the validation set were annotated by a group of medical experts, highlighting the region with the disease using a bounding box. Given the aforementioned annotations containing a bounding box $\mathbf{b} \in \{0, 1\}^{p}$ where $\mathbf{b}_i = 1$ denotes the regions within the bounding box and $\mathbf{b}_i = 0$ outside and an interpretability map $\mathbf{m} \in \mathbb{R}^{p}$ obtained via one of the methods described in Section~\ref{sec:int_methods}, we evaluate the correctness of interpretability maps using two metrics: pointing game and Intersection over union (IoU).

\textbf{Pointing Game}\,\textendash\,
Proposed by~\cite{zhang2018top}, the pointing game measures the precision of interpretability methods~\cite{jung2021explaining}. It finds the position in the map with the highest activation $a = \arg \max (\mathbf{m})$. If $\mathbf{b}_{a} = 1$, it is a "Hit". Otherwise, it is a "Miss". Based on this, we can calculate the average pointing game accuracy by measuring $\text{Acc} = \frac{\#\text{Hit}}{\#\text{Hit} + \#\text{Miss}}$.

\textbf{Intersection over union}\,\textendash\,When the bounding box surrounding the area of interest is large, such as in the Kvasir dataset, interpretability methods may appear inflated in performance using the pointing game. To conduct a more accurate investigation, we employ IoU metric. 

Based on $\mathbf{m}$, we select the pixels with activations higher than the top $k$th percentile ($p_k$) and produce a binary interpretability map denoted as $\bm{t}$ where $\bm{t}_i =
\begin{cases}
1 & \text{if} \quad \bm{m}_i \geq p_k \\
0,              & \text{otherwise}
\end{cases}
$. In this work, we use $k=5$. Then, we estimate the tightest bounding box around $\bm{t}$ where $\bm{t}_i = 1$. Given the bounding box estimation for the interpretability map $\bm{t}$ and the ground truth $\bm{b}$, we calculate $\text{IoU} = \frac{\bm{t}\,\cap\,\bm{b}}{\bm{t}\,\cup\,\bm{b}}$. Visual summary of both the pointing game and the IoU overlap are provided in \figurename~\ref{fig:vis_example}.

\begin{table}[t]
\centering
\scriptsize
\caption{Performance of ViT-B/16 models trained on datasets from Section~\ref{sec:dataset} with random, supervised, and self-supervised initializations. Validation accuracies exceeding the benchmark are highlighted in bold and * denotes approximate benchmark accuracy.}
\label{tbl:training_out}
\begin{tabular}{ccccc|c}
\toprule
\multirow{2}{*}{Dataset} & \multirow{2}{*}{\shortstack{Model\\initialization}} & \multicolumn{3}{c}{\phantom{----}Validation} & \multirow{3}{*}{\shortstack{Benchmark\\validation\\accuracy}} \\
\cmidrule[0.75pt]{3-5}
~ & ~ & \phantom{--}Accuracy\phantom{--} & \phantom{--}Precision\phantom{--} & \phantom{--}Recall\phantom{--} & ~ \\
\midrule

\multirow{4}{*}{CP-Child} & Random & 96.00 & 92.36 & 88.66 & ~ \\
~ & Supervised & \textbf{99.71} & 99.00 & 99.66 & \multirow{2}{*}{\shortstack{99.29~\cite{wang2020improved}}} \\
~ & DINO & \textbf{99.64} & 99.00 & 99.33 & ~ \\
~ & MAE & \textbf{99.57} & 99.33 & 98.66 & ~ \\
\cmidrule[0.75pt]{1-6}
\multirow{4}{*}{DUKE} & Random & 74.63 & 70.19 & 86.57 & ~ \\
~ & Supervised & \textbf{78.88} & 75.60 & 85.96 & \multirow{2}{*}{\shortstack{ 76.00*~\cite{duke_intrinsic}}} \\
~ & DINO & \textbf{78.53} & 79.48 & 77.55 & ~ \\
~ & MAE & \textbf{79.78} & 77.26 & 85.02 & ~ \\
\cmidrule[0.75pt]{1-6}
\multirow{4}{*}{Kvasir} & Random & 69.25 & 73.05 & 61.00 & ~ \\
~ & Supervised & \textbf{85.25} & 89.38 & 80.00 & \multirow{2}{*}{\shortstack{ 77.00*~\cite{kvasir}}} \\
~ & DINO & \textbf{85.00} & 87.63 & 81.50 & ~ \\
~ & MAE & \textbf{85.50} & 90.34 & 79.50 & ~ \\
\cmidrule[0.75pt]{1-6}
\multirow{4}{*}{MURA} & Random & 70.15 & 78.68 & 51.63 & ~ \\
~ & Supervised & 80.63 & 84.90 & 72.41 & \multirow{2}{*}{\shortstack{81.20~\cite{mura_benchmark}}} \\
~ & DINO & 81.01 & 86.65 & 71.30 & ~\\
~ & MAE & \textbf{82.23} & 85.21 & 76.07 & ~ \\
\bottomrule
\end{tabular}
\end{table}

\section{Experimental Results and Discussion}
\label{sec:experimental_results}

Using a grid-search approach for finding the most-suitable hyperparameters, we train four models on the datasets described in Section~\ref{sec:dataset} and present results in Table~\ref{tbl:training_out}. Compared to the benchmark results from various studies, our models demonstrate results that are comparable to the state-of-the-art, thus making them suitable for experiments on explainability. Further details on the employed training approach and the final models can be found in this repository: \url{https://github.com/ugent-korea/attention_maps}.

\begin{figure}[t!]
\centering
\begin{subfigure}{0.45\textwidth}
\centering
\includegraphics[width=0.32\textwidth]{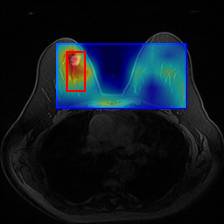}
\includegraphics[width=0.32\textwidth]{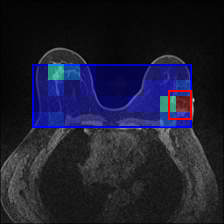}
\includegraphics[width=0.32\textwidth]{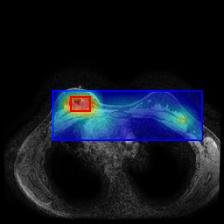}
\includegraphics[width=0.32\textwidth]{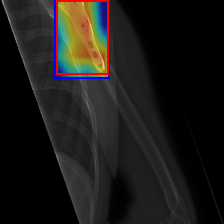}
\includegraphics[width=0.32\textwidth]{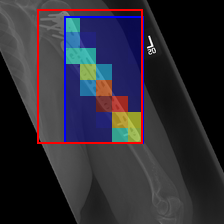}
\includegraphics[width=0.32\textwidth]{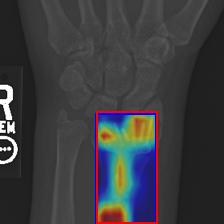}
\includegraphics[width=0.32\textwidth]{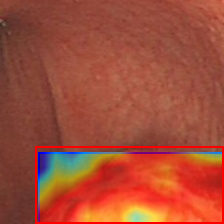}
\includegraphics[width=0.32\textwidth]{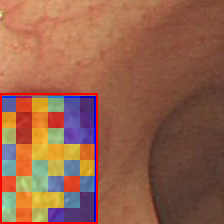}
\includegraphics[width=0.32\textwidth]{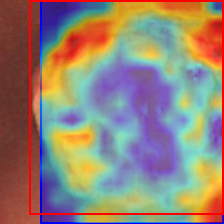}
\includegraphics[width=0.32\textwidth]{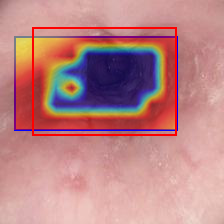}
\includegraphics[width=0.32\textwidth]{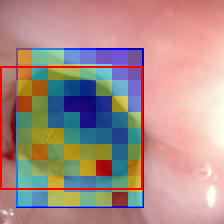}
\includegraphics[width=0.32\textwidth]{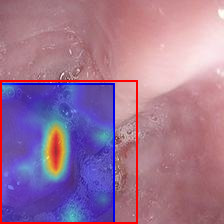}
GradCAM \phantom{--.} Attention \phantom{-----} Chefer \phantom{--} \\
\caption{Accurate interpretability maps with high IoU scores}
\label{fig:gradcam_good}
\end{subfigure}
\hspace{0.2em}
\begin{subfigure}{0.45\textwidth}
\centering
\includegraphics[width=0.32\textwidth]{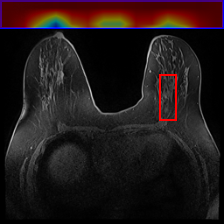}
\includegraphics[width=0.32\textwidth]{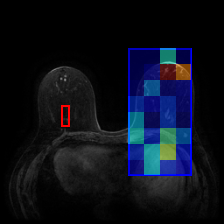}
\includegraphics[width=0.32\textwidth]{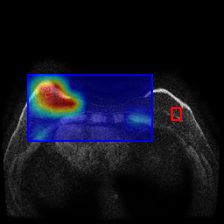}
\includegraphics[width=0.32\textwidth]{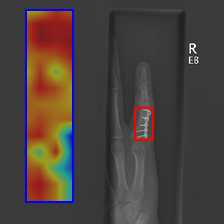}
\includegraphics[width=0.32\textwidth]{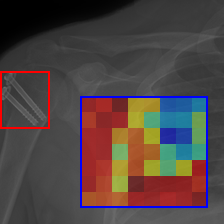}
\includegraphics[width=0.32\textwidth]{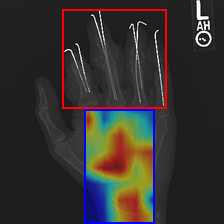}
\includegraphics[width=0.32\textwidth]{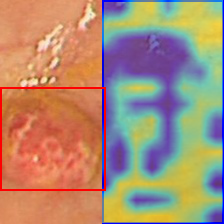}
\includegraphics[width=0.32\textwidth]{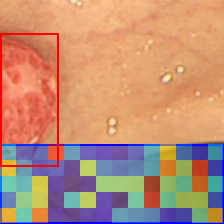}
\includegraphics[width=0.32\textwidth]{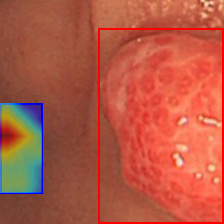}
\includegraphics[width=0.32\textwidth]{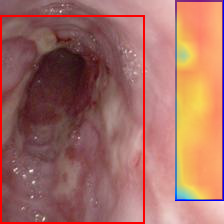}
\includegraphics[width=0.32\textwidth]{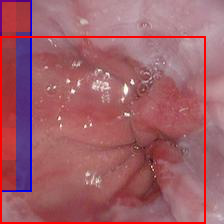}
\includegraphics[width=0.32\textwidth]{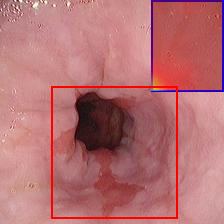}
GradCAM \phantom{--.} Attention \phantom{-----} Chefer \phantom{--} \\
\caption{Inaccurate interpretability maps with low or zero IoU scores}
\label{fig:gradcam_bad}
\end{subfigure}
\caption{Qualitative examples generated using interpretability methods from Section~\ref{sec:int_methods} on the four datasets employed in this study. Red boxes highlight annotations made by medical experts, whereas blue boxes indicate regions with intensity levels exceeding the top 5\%, as identified by the interpretability methods. When the red and blue areas overlap, it indicates a high IoU score, whereas low overlap indicates a low score. When the two boxes do not intersect, it means that the IoU is zero.}
\label{fig:vis_qualitative}
\end{figure}

\begin{figure}[t!]
\centering
\begin{subfigure}{0.48\textwidth}
\centering
\includegraphics[width=1\textwidth]{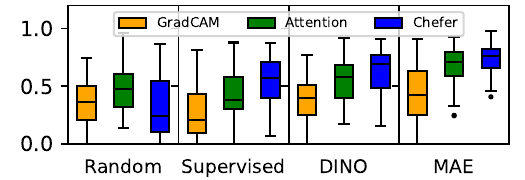}
\caption{CP-Child}
\label{fig:cpchild_boxplot}
\end{subfigure}
\begin{subfigure}{0.48\textwidth}
\centering
\includegraphics[width=1\textwidth]{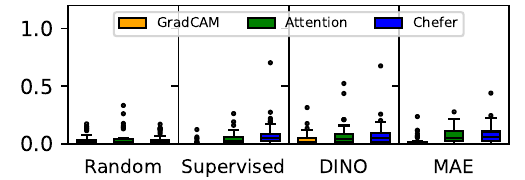}
\caption{DUKE}
\label{fig:duke_boxplot}
\end{subfigure}
\begin{subfigure}{0.48\textwidth}
\centering
\includegraphics[width=1\textwidth]{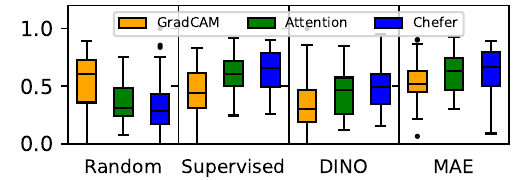}
\caption{Kvasir}
\label{fig:kvasir_boxplot}
\end{subfigure}
\begin{subfigure}{0.48\textwidth}
\centering
\includegraphics[width=1\textwidth]{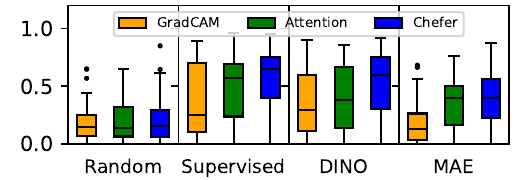}
\caption{MURA}
\label{fig:mura_boxplot}
\end{subfigure}
\caption{Box plots showing the IoUs of evaluated interpretability methods for datasets from Section~\ref{sec:dataset}. Each panel illustrates the distribution of IoU values across four pretraining strategies: Random, Supervised, DINO, and MAE.}
\label{fig:boxplot_images}
\end{figure}

Using the trained models with state-of-the-art performance, we follow the protocol detailed in Section~\ref{sec:evaluation} and present experimental results on interpretability in \figurename~\ref{fig:vis_qualitative}, \figurename~\ref{fig:boxplot_images}, and Table~\ref{tbl:vis_iou_out}. In particular, \figurename~\ref{fig:vis_qualitative} illustrates several qualitative examples, while Table~\ref{tbl:vis_iou_out} displays pointing game accuracy and mean IoU scores. Finally, \figurename~\ref{fig:boxplot_images} represents IoU distributions represented in the form of boxplots. Based on these results, we make the observations below.

\textbf{Qualitative results can be misleading}.
In \figurename~\ref{fig:vis_qualitative}, we present qualitative interpretability outputs for all methods considered. As can be seen, depending on the selected subset of images and the method, both accurate and inaccurate interpretability maps can be found. These findings emphasize the danger of making strong claims about interpretability methods based solely on qualitative results, including attention maps.

\textbf{Pretraining method influences the interpretability outcome}.
While most models achieve similar accuracies on the same dataset, their interpretability outcomes differ significantly, as seen in Table~\ref{tbl:vis_iou_out}. This indicates that pretraining methods indeed affect interpretability outcomes. Specifically, in the majority of cases (though not all), self-supervised pretrained models yield higher performance with attention maps. However, contrasting the findings reported by~\cite{dino}, we find that interpretability results from DINO may not necessarily outperform those from supervised or MAE pretraining. This suggests that certain pretraining tasks may be more suitable for specific medical datasets than others.

\begin{table}[t!]
\centering
\scriptsize
\caption{Comparison of interpretability methods evaluated using the pointing game and IoU overlap for various ViT initializations across four datasets. The largest pointing game accuracy for each row is highlighted in bold font, while the highest average IoU overlap is underlined.}
\label{tbl:vis_iou_out}
\begin{tabular}{ccccc|ccc}
\toprule
\multirow{2}{*}{Dataset} & \multirow{2}{*}{\shortstack{Model\\init.}} & \multicolumn{3}{c}{Pointing game} & \multicolumn{3}{c}{IoU} \\
\cmidrule[0.75pt]{3-8}
~ & ~ & GradCAM & Attention & Chefer & GradCAM & Attention & Chefer \\
\midrule
\multirow{4}{*}{CP-Child} & Random & 0.64 & \textbf{0.96} & 0.54 & 0.36 & \underline{0.45} & 0.33 \\
~ & Supervised & 0.38 & 0.36 & \textbf{0.54} & 0.27 & 0.42 & \underline{0.55} \\
~ & DINO & 0.76 & 0.68 & \textbf{0.86} & 0.38 & 0.54 & \underline{0.63} \\
~ & MAE & 0.56 & 0.96 & \textbf{0.98} & 0.41 & 0.67 & \underline{0.73} \\
\midrule
\multirow{4}{*}{DUKE} & Random & 0.04 & \textbf{0.18} & 0.12 & 0.03 & \underline{0.04} & 0.03 \\
~ & Supervised & 0.00 & \textbf{0.30} & 0.28 & 0.00 & 0.04 & \underline{0.08} \\
~ & DINO & 0.10 & \textbf{0.52} & 0.40 & 0.03 & 0.07 & \underline{0.08} \\
~ & MAE & 0.12 & \textbf{0.36} & \textbf{0.36} & 0.01 & 0.07 & \underline{0.08} \\
\midrule
\multirow{4}{*}{Kvasir} & Random & 0.88 & \textbf{0.98} & 0.72 & \underline{0.54} & 0.36 & 0.34 \\
~ & Supervised & 0.80 & 0.88 & \textbf{0.92} & 0.44 & 0.59 & \underline{0.61} \\
~ & DINO & 0.72 & \textbf{0.96} & 0.90 & 0.32 & 0.43 & \underline{0.48} \\
~ & MAE & 0.52 & \textbf{0.82} & 0.80 & 0.52 & 0.59 & \underline{0.62} \\
\midrule
\multirow{4}{*}{MURA} & Random & 0.30 & \textbf{0.40} & 0.34 & 0.18 & 0.19 & \underline{0.20} \\
~ & Supervised & 0.56 & \textbf{0.94} & \textbf{0.94} & 0.36 & 0.47 & \underline{0.56} \\
~ & DINO & 0.78 & \textbf{0.90} & 0.86 & 0.35 & 0.41 & \underline{0.52} \\
~ & MAE & 0.30 & \textbf{0.76} & 0.68 & 0.18 & 0.35 & \underline{0.39} \\
\bottomrule
\end{tabular}
\end{table}

\textbf{GradCAM interpretability for ViTs is inadequate}.
In both evaluation types and across multiple models, GradCAM is significantly outperformed by both attention maps and Chefer method, revealing that both methods are more appropriate for ViT interpretability compared to GradCAM, which was originally proposed for CNNs.

\textbf{Attention maps show promise for interpretability}.
As shown in Table~\ref{tbl:vis_iou_out} and Figure~\ref{fig:boxplot_images}, interpretability results of attention maps and the Chefer method are comparable. Attention maps generally perform better in the pointing game, while the Chefer method yields better IoU results. These observations hold true for the majority of models across all datasets, with the exception of CP-Child, where the Chefer method outperforms attention maps in the pointing game. Based on these findings, we suggest that researchers employ attention maps when the goal is to identify the most important location in an image. In other scenarios, however, we recommend using the Chefer method. In summary, while attention maps show promise, we find that the Chefer method is a more appropriate choice for interpretability in medical datasets.

\textbf{Bounding box annotations may be inadequate for interpretability evaluation on medical datasets}.
In this work, we worked with expert clinicians to highlight disease-positive regions with a bounding box to evaluate the efficacy of interpretability maps. However, we discover that this approach comes with significant shortcomings. For instance, in the MURA and Kvasir datasets, regions of interest often occupy large spaces, resulting in annotation boxes that include non-target areas, and leading to inflated results regardless of the method's actual precision. Therefore, using segmentation maps, which provide detailed pixel-level annotations of the areas of interest, could offer a more precise evaluation in future research.

\vspace{-1em}
\section{Conclusions and Future Perspectives}
AI-assisted medical imaging holds significant potential to revolutionize medical diagnostic processes. Despite the superb performance of ViTs on various medical imaging tasks, widespread adoption of these models has been hindered by explainability issues and the opaque nature of their decision-making processes. Recently, the use of attention maps as a method of explainability has garnered significant research interest, as attention maps are an integral part of the ViT’s decision-making process and directly influence its predictions. In this work, we investigated the explainability performance of attention maps and compared them to two commonly used methods: GradCAM and Chefer method. Using state-of-the-art ViT-B/16 models pretrained through various methods and applied to four medical datasets involving different imaging modalities and body parts, we found that, while attention maps show promise, the Chefer method should be preferred for explainability on medical datasets thanks to its superior results.

Throughout our investigation, we also discovered that commonly used interpretability evaluation methods relying on bounding box information have significant shortcomings in the context of medical imaging. As such, we advise future research efforts to refrain from using this type of evaluation and instead employ segmentation maps for a more precise assessment.

In this work, we evaluated the quality of visual explanations based on the annotations of experts. Recent research in this field suggests that this type of evaluation might be misleading in identifying the best-performing explainability method due to several reasons, with spurious correlations being the primary one~\cite{baniecki2023careful}. As such, we suggest that practitioners exercise caution when selecting methods and conduct thorough analyses to assess each method's effectiveness and reliability, considering potential biases and limitations in the evaluation process.

\bibliographystyle{splncs04}
\bibliography{main}

\end{document}